# Optimizing Fire Safety: Reducing False Alarms Using Advanced Machine Learning Techniques

Muhammad Hassan Jamal[1], Abdulwahab Alazeb[2],
Shahid Allah Bakhsh[3], Wadii Boulila[4], Syed Aziz Shah[5],
Aizaz Ahmad Khattak[6], Muhammad Shahbaz Khan[6*]

[1]Department of Computer Sciences, Quaid-i-Azam University, Islamabad, 45320, Pakistan.
[2]Department of Computer Science, College of Computer Science and Information Systems, Najran University, Najran, 61441, Saudi Arabia.
[3]Department of Cyber Security, National University of Science & Technology, Karachi, 75350, Pakistan.
[4]College of Computer and Information Sciences, Prince Sultan University, Riyadh, 11586, Saudi Arabia.
[5]Centre for Intelligent Healthcare, Coventry University, Coventry, CV1 5FB, U.K..
[6*]School of Computing, Engineering and the Built Environment, Edinburgh Napier University, Edinburgh, EH10 5DT, U.K..

*Corresponding author(s). E-mail(s):
muhammadshahbaz.khan@napier.ac.uk;
Contributing authors: mhassan@cs.qau.edu.pk; afalazeb@nu.edu.sa;
shahid.mscys21pnec@student.nust.edu.pk; wboulila@psu.edu.sa;
syed.shah@coventry.ac.uk; 40614576@live.napier.ac.uk;

**Abstract**

Fire safety practices are important to reduce the extent of destruction caused by fire. While smoke alarms help save lives, firefighters struggle with the increasing number of false alarms. This paper presents a precise and efficient Weighted ensemble model for decreasing false alarms. It estimates the density, computes weights according to the high and low-density regions, forwards the high region weights to KNN and low region weights to XGBoost and combines the predictions. The proposed model is effective at reducing response time, increasing fire safety, and minimizing the damage that fires cause. A specifically designed dataset for smoke detection is utilized to test the proposed model. In addition, a variety of ML models, such as Logistic Regression (LR), Decision Tree (DT), Random Forest (RF), Naïve Bayes (NB), K-Nearest Neighbour (KNN), Support Vector Machine (SVM), Extreme Gradient Boosting (XGBoost), Adaptive Boosting (ADAB), have also been utilized. To maximize the use of the smoke detection dataset, all the algorithms utilize the SMOTE re-sampling technique. After evaluating the assessment criteria, this paper presents a concise summary of the comprehensive findings obtained by comparing the outcomes of all models.

# 1 Introduction

A fire is a highly destructive incident that can be encountered by any individual. Smoke detectors play a crucial role in the quick and accurate detection of fires, ultimately leading to the survival of numerous lives [1]. The increased emphasis on fire safety regulations and the extensive adoption of smoke detectors significantly contribute to the reduction in fire incidents. Smoke alarms are present in 96% of households in the United States. However, 20% of these homes contain non-functional smoke alarms [2]. In addition, most of the time traditional smoke detectors generate false alarms. The increasing occurrence of erroneous fire alarms is a significant challenge for firefighters.

In particular, the two most significant types of smoke detectors are photoelectric and ionization smoke detectors. Photoelectric smoke detectors include photoelectric receivers, lenses, and light sources that emit infrared, visible, or ultraviolet light [3]. The photosensor occasionally picks up light that the light source has emitted and that has traveled through the air during testing. These detectors are also known as optical detectors [4]. Whereas in Ionization smoke detectors, a radioisotope charges the air with ions. If any ions reach the open chamber, they will adhere to the smoke particles, making them unable to conduct the current. An electrical circuit activates the alarm when it detects a voltage difference between the two compartments [5]. Reducing false alarms can lead to prompt action in the event of a real fire emergency. It increases the public's responsiveness to real fire alarms, reducing the risk of fatalities. It also curtails the risk of injury for both the public and firefighters. Additionally, it can reduce unnecessary disruptions and productivity losses, thereby enhancing the effectiveness, efficiency, profitability, and services of businesses and organizations [6]. The researchers are more focused on reducing false alarms using Machine Learning (ML) techniques. Machine learning and deep learning methods are extensively being utilized for smoke and fire detection.

This study examines the results of different ML models used in the smoke detection dataset to reduce the occurrence of false alarms. These models exhibited high accuracy and efficiency. We employ various ML models, such as logistic regression (LR), to perform binary classification tasks by generating probability-based predictions. Next, we explore nonparametric decision trees (DTs) and Random Forests (RFs), which effectively handle complex feature-target relationships. We also use Naive Bayes, the k-nearest neighbour (KNN), and the support vector machine (SVM) due to their adaptability and effectiveness in classification applications. The XGBoost and ADAB are utilized, which are boosting algorithms that employ an ensemble technique to combine multiple weak learners into a single strong one. Additionally, we have used the Proposed Weighted ensemble model, an ensemble technique based on local density. The dataset examines a smoke detector that uses AI sensor fusion to determine whether a fire alarm is triggered. The dataset includes a "Fire Alarm" goal variable containing binary-categorized tagged data. Figure 1 illustrates the relationship between several characteristics and the goal variable in the dataset. A correlation coefficient visually displays the strength and direction of the relationship between the targeted variable and each attribute. To build reliable models, it is crucial to select suitable independent variables. Predicting the correlation coefficients allows us to identify the most valuable traits for target variable predictions. When creating predictions, assigning greater importance to features with high correlation coefficients is common.

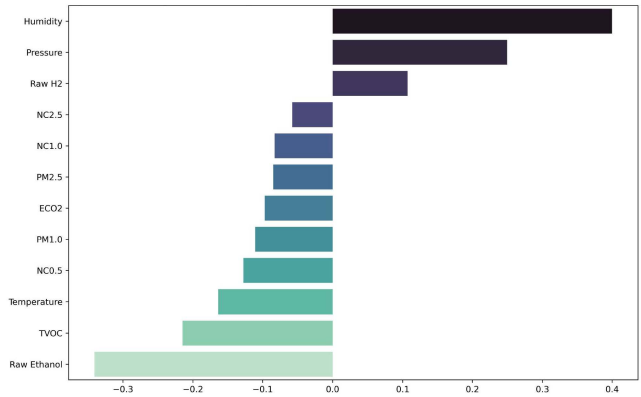

**Fig. 1** Correlation coefficient between features and Fire Alarm.

The main contributions of this article are:

- The Weighted ensemble model has been implemented, which calculates the density and divides it into high and low regions. The high and low region weights were passed to the KNN and XGBoost, respectively, and their predictions were combined at the end.
- The state-of-the-art ML models were applied to the smoke detection dataset to compare the results with the suggested Weighted ensemble model and comparative analysis has been provided.

## 2 Methodology

This section provides a concise overview of the procedures undertaken throughout the experiment. Before analyzing the study's results, we pre-process, train, and test the data. The analysis provided a comprehensive breakdown of multiple evaluation criteria. Figure 2 illustrates the entire procedure's layout.

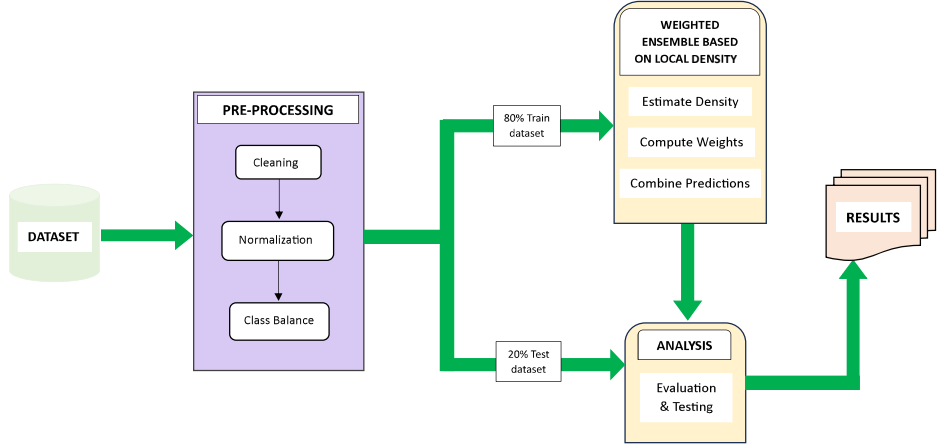

**Fig. 2** Methodology for the analysis of smoke detection dataset.

## 2.1 Dataset

The dataset we used is the Smoke detection dataset which is publicly available on Kaggle [7], [8]. The dataset uses Internet of Things (IoT) devices to gather training data to develop an AI-powered smoke detector system. To create a high-quality training dataset, it is necessary to sample from a diverse range of ecosystems and fire sources [9]. Here is a concise summary of the different captured scenarios: Normal indoor and outdoor features include a standard firefighter training area, an indoor gas fireplace, an indoor wood fireplace, an outdoor grill that uses coal, gas, or wood, and an outdoor area with high humidity. The collection contains approximately sixty thousand readings. Each sensor operates at a sample rate of 1 Hz. The dataset has a Coordinated Universal Time (UTC) timestamp to ensure data monitoring of every sensor reading. The dataset features are highlighted in Table 1.

## 2.2 Pre-Processing

Several steps have been followed to conduct pre-processing on the smoke detection dataset. Data cleaning eliminates any potential for ambiguity or inaccuracy in your data. Each dataset consists of a substantial number of elements. Before training a model, it is advisable to pre-process the dataset by eliminating any possibly erroneous data. The objective of normalization is to ensure that the features are standardized and coherent, thereby enhancing the prediction capabilities of the model. The ML model's training process becomes more simplified when the inputs are equalized [10]. A dataset with a substantial number of records is the foundation for the experiment's model. This dataset contains a statistically significant disparity between the classes labeled "Alarm" and "No Alarm," which consist of 44,757 and 17,873 instances, respectively, as illustrated in Figure 3. Before training and testing the model, ensuring that each class has an equal number of instances is crucial. Techniques such as oversampling and undersampling can help achieve balance. We employed SMOTE, an oversampling method, to equalize the proportion of the minority class with that of the majority class [11]. To achieve this goal, SMOTE produces artificial samples along the lines in the feature space that link the nearest neighbours. Figure 4 demonstrates that both classes have an equal number of instances, specifically 44757 instances each, after applying SMOTE. In experiments, the study effectively addresses the overfitting problem by utilizing an 80/20 train-test split. Sklearn library employed a train-test-split approach to ensure precise data splitting.

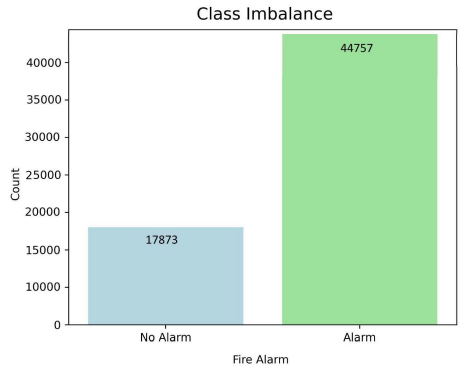

**Fig. 3** Classes of Smoke detection dataset before using SMOTE technique.

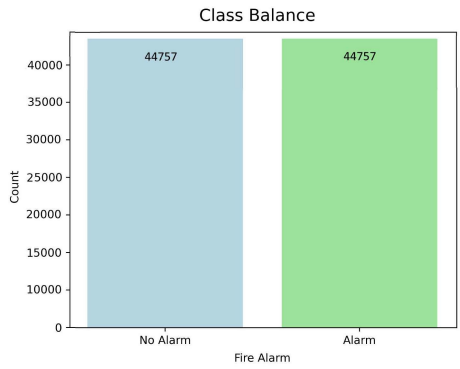

**Fig. 4** Classes of Smoke detection dataset after using SMOTE technique.

**Table 1** Smoke detection dataset features.

| Feature | Name |
| --- | --- |
| Humidity | Air humidity in percentage (%) |
| Pressure | Air pressure in hectoPascals (hPa) |
| Temperature | Air temperature in Celsius (C) |
| CNT | Sample counter |
| eCO2 | CO2 equivalent concentration in parts per million (ppm) |
| Fire Alarm | If there is a fire, Ground truth is "1" |
| NC 0.5, NC 1.0 and NC 2.5 | The number concentration known as NC quantifies the exact number of particles present in the air, in contrast to PM |
| PM 1.0 and PM 2.5 | Size of particulate matter |
| Raw Ethanol | Raw ethanol gas |
| Raw H2 | Raw molecular hydrogen |
| TVOC | The total number of volatile organic compounds per billion parts (ppb) |
| UTC | Coordinated Universal Time in seconds |

## 2.3 The Proposed Weighted Ensemble Model Based on Local Density

The proposed method, as explained in Figure 5, combines KNN and XGBoost predictions based on local density for every data point. The process starts by collecting data from different IoT devices. We pre-process and resample the data by following the above-mentioned procedure. We then estimate the density using the nearest neighbour algorithm. We divide the density into two regions: high density and low density. We forward the high-density region to KNN, which assigns higher weights due to its ability to capture local patterns. Furthermore, we pass the low-density region to XGBoost, which captures global patterns and secures higher weights from the region. We computed and combined the final predictions by multiplying each model's prediction by its weight. We then analyzed the predictions and obtained the results at the end.

## 2.4 Evaluation

The accuracy of a model, which refers to the percentage of correct predictions it makes, is the primary statistic used for evaluation. We calculate the model's accuracy by dividing the total number of correct forecasts by the total number of predictions made by the model. Precision measures the entire number of positive outcomes, both true and false, divided by the total number of positive outcomes to get the calculation. Recall is the sum of true positives divided by the sum of false negatives, which should be used to ascertain the accuracy rate. The F-1 score requires assigning equal weights to recall and precision in the average. A confusion matrix displays the frequencies of correct, incorrect, true negative, and false negative predictions for a specific model in a specific format. The goal is to evaluate the efficacy of a model in a binary classification task. The Area Under the Curve (AUC) determines the power of a binary classification model concerning positive and negative values. The AUC ranges are from 0 to 1, where 1 represents an optimal classifier and 0.5 represents a random classifier [12].

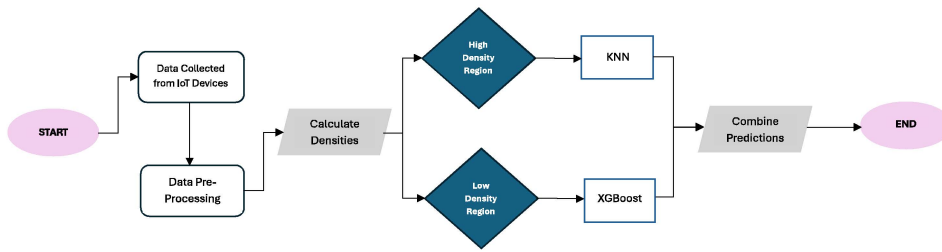

**Fig. 5** Flowchart of the Proposed Weighted Ensemble Model Based on Local Density.

# 3 Results and Analysis

ML systems test the accuracy of smoke detection outcome predictions in this study. An in-depth analysis of the outcomes produced by various ML models sheds light on their strengths and applicability to specific problems.

## 3.1 Weighted Ensemble Model

The training data was processed and organized during each stage of the model's development. A comprehensive experimental setup has been developed, and the outcomes have been utilized to assess the model comparison. After evaluating the assessment criteria, we present a concise summary of the comprehensive findings obtained by comparing the outcomes of the proposed and state-of-the-art models. We evaluated and compared the models' performance using evaluation metrics. This study employed the SMOTE to assess and contrast the model's output. To accomplish the study objective, we compared the test results among models. Subsequently, the models' performance was compared using the smoke detection dataset.

The Proposed Weighted ensemble model has performed efficiently on the smoke detection dataset, as mentioned in Table 2. The model has gained an accuracy of 0.999832 in less time when compared with other models. It also has a precision of 0.999789, a recall of 0.999889, an F-1 score of 0.999833, and an AUC score of 0.999829. We have also generalized the Weighted ensemble model for various real-world scenarios, utilizing the reliability parameters of Cohen's Kappa and Matthews' correlation coefficient. The Cohen's Kappa Coefficient achieved a value of 0.999495, which is optimal as the range for this coefficient (0.61 to 0.80) is substantial and (0.81 to 1.00) is an optimal range [13]. The Matthews' Correlation Coefficient range (0.40 to 0.69) indicates a random prediction, while (0.70 to 1.00) indicates a perfect prediction, and our model scored 0.999665. Figure 6 depicts the proposed model's confusion matrix, which shows the number of actual and predicted values that are true and false. The value 8903 represents the truly predicted actual values, while the values 2 and 1 represent the total of false positives and negatives, respectively. Lastly, the value 8997 depicts the true negatives of the Proposed Weighted ensemble model.

**Table 2** Results of Proposed Weighted Ensemble Model.

| Evaluation Parameters | Model Results |
| --- | --- |
| Accuracy | 0.999832 |
| Precision | 0.999789 |
| Recall | 0.999889 |
| F-1 score | 0.999833 |
| AUC | 0.999829 |
| Cohen's Kappa Coefficient | 0.999495 |
| Matthews' Correlation Coefficient | 0.999665 |

Fig. 6 Confusion Matrix of the Proposed Weighted Ensemble Model.

## 3.2 Comparative Analysis

The Logistic Regression algorithm, a supervised machine learning method, calculates the probability of an instance belonging to a specific class. The Decision Trees are built by iteratively partitioning the data into increasingly smaller pieces. Each division optimizes the information gained by splitting the data based on a specified attribute. The Random Forest method calculates the average of all the outputs for regression and classification tasks or makes a collective decision based on voting during prediction. Adaptive Boosting, a boosting technique, utilizes the step-wise addition strategy to generate a limited number of strong learners by using multiple weak learners. Extreme Gradient Boosting constructs an ensemble model using a boosting technique in which each succeeding weak learner corrects the mistakes made by its predecessors [14]. The Naïve Bayes classifier estimates the likelihood of an instance belonging to a particular class by analyzing feature values. The K-Nearest Neighbour method identifies the KNNs at a given data point. Afterward, it utilizes an average or majority vote from the K neighbours to determine the class or value of the data item. The support vector machine selects the optimal N-dimensional hyperplane for classifying the feature space. The purpose of the hyperplane is to maximize the distance between neighbouring data points that belong to different categories. [15].

The experimental results indicate that the performance of DT surpasses that of the other models in terms of accuracy. The Weighted ensemble model and RF outperform other ML models. The RF, KNN, and Weighted ensemble models outperform others in terms of precision scores. Additionally, the RF and Weighted ensemble models excel at recall scores. The XGBoost model achieves a high F-1 score, while the SVM model has superior performance in terms of AUC score compared to the other models that have performed below par. Lastly, the Weighted ensemble model excels in both areas under the curve (AUC) and recall score. Table 3 displays the comparison results between the models built using the SMOTE technique. The Weighted ensemble, RF, and KNN models exhibited superior performance. Upon evaluating the NB model, its

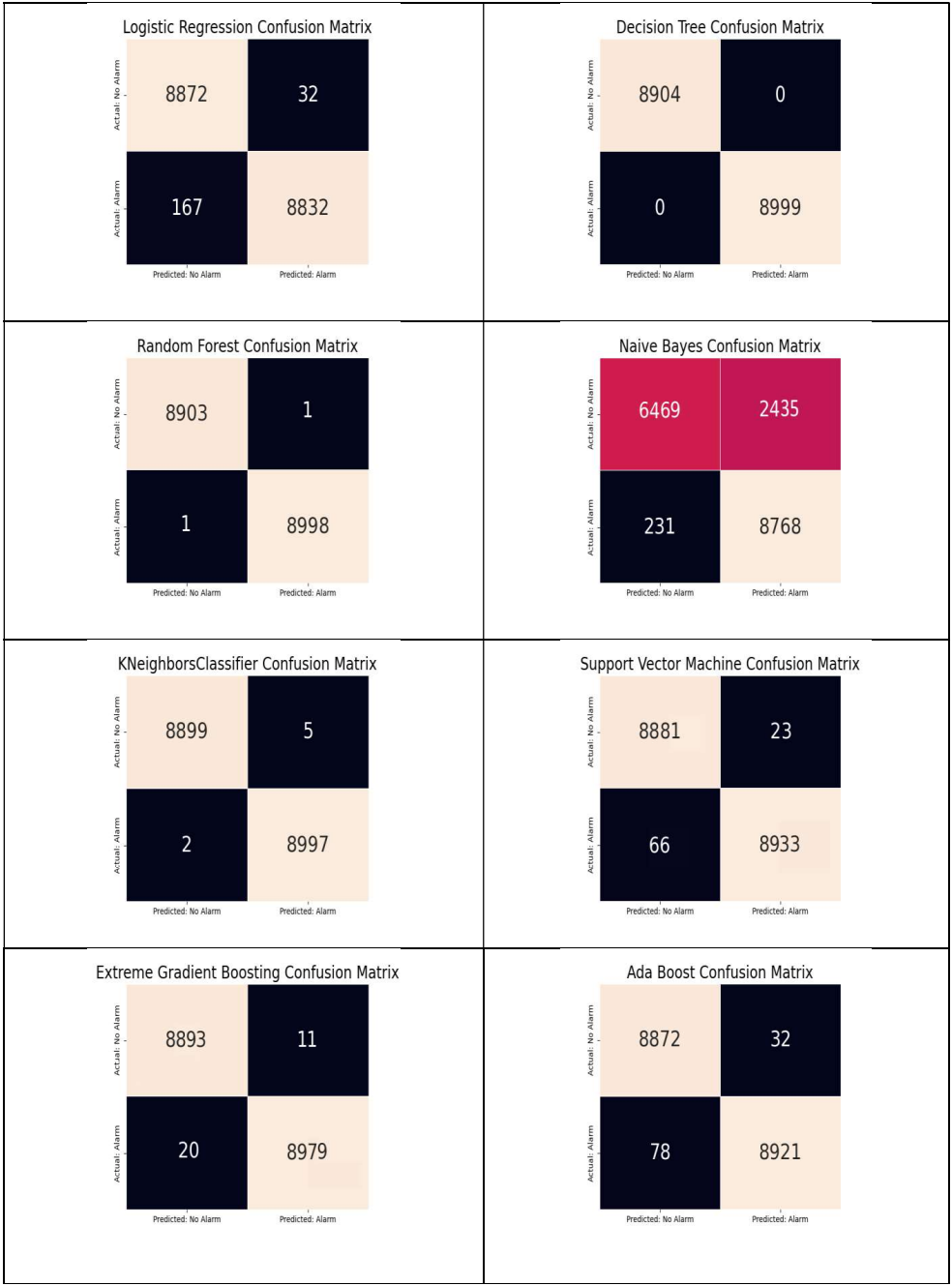

**Fig. 7** Confusion Matrices of ML Models.

Table 3 Comparison of ML models utilized on smoke detection dataset.

| Model | Accuracy | Precision | Recall | F-1 Score | AUC |
|---|---|---|---|---|---|
| Logistic Regression | 0.988885 | 0.981442 | 0.996389 | 0.988860 | 0.988924 |
| Decision Tree | 1.000000 | 1.000000 | 1.000000 | 1.000000 | 1.000000 |
| Random Forest | 0.999888 | 0.999888 | 0.999888 | 0.999887 | 0.998888 |
| Naïve Bayes | 0.851086 | 0.974330 | 0.782647 | 0.868032 | 0.825753 |
| K-Nearest Neighbour | 0.999609 | 0.999778 | 0.999444 | 0.999611 | 0.999608 |
| Support Vector Machine | 0.995028 | 0.992665 | 0.997431 | 0.995042 | 0.994961 |
| Extreme Gradient Boosting | 0.998268 | 0.997777 | 0.998776 | 0.998276 | 0.998945 |
| Adaptive Boosting | 0.993856 | 0.991332 | 0.996425 | 0.993873 | 0.993869 |
| **Weighted Ensemble Model** | **0.999832** | **0.999789** | **0.999889** | **0.999833** | **0.999829** |

accuracy is 0.851086, precision is 0.974330, recall is 0.782647, F-1 score is 0.868032, and AUC is 0.8257530. These metrics collectively indicate that the NB model has the lowest performance among the models evaluated. Although the Weighted ensemble and RF models exhibited higher accuracy, the KNN, LR, ADAB, and SVM models also showed commendable results. When considering total accuracy, the DT algorithm achieves a perfect score of 1.000000, followed closely by RF with a score of 0.999888, the Weighted ensemble model with a score of 0.999832, KNN with a score of 0.999609, XGBoost with a score of 0.998268, SVM with a score of 0.995028, ADAB with a score of 0.993856, and LR with a score of 0.988885.

According to the data obtained from the smoke detection dataset, the DT model outperforms all other techniques. Based on the findings, it is evident that the Weighted ensemble model and RF rank as the second-most effective models for this study. Several other models had slightly diminished accuracy and precision overall. The confusion matrices have been displayed in Figure 7, which yields the tally of accurate, inaccurate, true negative, and false negative estimations of all the models employed in this study.

## 4 Conclusion

The integration of machine learning (ML) with fire detection has yielded significant outcomes, precipitating advancements in fire protection methodologies. The Proposed Weighted ensemble model detects anomalies, allowing prompt rescue operations and efficient resource distribution, reducing the number of victims, and minimizing damage. To identify the regions with the highest probability of experiencing fires, the proposed model can analyze many data elements, such as weather conditions, building materials, and historical fire patterns. Moreover, the Proposed Weighted ensemble model can acquire knowledge and continuously adjust to novel data. ML has significantly transformed fire protection methods. In addition to mitigating property damage and preserving lives, it has equipped individuals and authorities with the necessary information to avert fires proactively. Consequently, fire detection systems can progressively enhance their performance over time. Because of their adaptability, the systems can detect emerging fire patterns and adjust their reactions accordingly. Various sensors strategically positioned in buildings and public spaces can offer instantaneous data that ML systems can analyze. As technology advances, we should expect the emergence of more advanced techniques for preventing and responding to fires. Integrating ML in fire detection has notably transformed our approach to fire protection, resulting in safer and more resilient communities.